# Bayesian structure learning using dynamic programming and MCMC

**Daniel Eaton and Kevin Murphy**
Computer Science Dept.
Universtity of British Columbia
Vancouver, BC
{deaton,murphyk}@cs.ubc.ca

## Abstract

MCMC methods for sampling from the space of DAGs can mix poorly due to the local nature of the proposals that are commonly used. It has been shown that sampling from the space of node orders yields better results [FK03, EW06]. Recently, Koivisto and Sood showed how one can analytically marginalize over orders using dynamic programming (DP) [KS04, Koi06]. Their method computes the exact marginal posterior edge probabilities, thus avoiding the need for MCMC. Unfortunately, there are four drawbacks to the DP technique: it can only use modular priors, it can only compute posteriors over modular features, it is difficult to compute a predictive density, and it takes exponential time and space. We show how to overcome the first three of these problems by using the DP algorithm as a proposal distribution for MCMC in DAG space. We show that this hybrid technique converges to the posterior faster than other methods, resulting in more accurate structure learning and higher predictive likelihoods on test data.

## 1 Introduction

Directed graphical models are useful for a variety of tasks, ranging from density estimation to scientific discovery. One of the key challenges is to learn the structure of these models from data. Often (e.g., in molecular biology) the sample size is quite small relative to the size of the hypothesis space. In such cases, the posterior over graph structures given data, $p(G|D)$, gives support to many possible models, and using a point estimate (such as MAP) could lead to unwarranted conclusions about the structure, as well as poor predictions about future data. It is therefore preferable to use Bayesian model averaging. If we are interested in the probability of some structural feature $f$ (e.g., $f(G) = 1$ if there is an edge from node $i$ to $j$ and $f(G) = 0$ otherwise), we can compute posterior mean estimate $E(f|D) = \sum_G f(G)p(G|D)$. Similarly, to predict future data, we can compute the posterior predictive distribution $p(x|D) = \sum_G p(x|G)p(G|D)$.

Since there are $O(d!2^{\binom{d}{2}})$ DAGs (directed acyclic graphs) on $d$ nodes [Rob73], exact Bayesian model averaging is intractable in general.[1] Two interesting cases are when it is tractable are averaging over trees [MJ06] and averaging over graphs given a known node ordering [DC04]. If the ordering is unknown, we can use MCMC techniques to sample orders, and sample DAGs given each such order [FK03, EW06, HZ05]. However, Koivisto and Sood [KS04, Koi06] showed that one can use dynamic programming (DP) to marginalize over orders analytically. This technique enables one to compute all marginal posterior edge probabilities, $p(G_{ij} = 1|D)$, exactly in $O(d2^d)$ time. Although exponential in $d$, this technique is quite practical for $d \leq 20$, and is much faster than comparable MCMC algorithms on similar sized problems[2]

Unfortunately, the DP method has three fundamental limitations, even for small domains. The first problem is that it can only be used with certain kinds of graph priors which satisfy a "modularity" condition, which we shall explain in Section 3 below. Although this seems like a minor technical problem, it can result in significant bias. This can lead to unwarranted conclusions about structure, even in the large sample setting, as we shall see below. The second problem is that it can only compute posteriors over modular features; thus it cannot be used to compute the probability of features like "is there a path between nodes $i$ and $j$ via $k$", or "is $i$ an ancestor of $j$". Such long-distance features are often of more interest than direct edges. The third problem is that it is expensive to compute predictive densi-ties, $p(x|D)$. Since the DP method integrates out the graph structures, it has to keep all the training data $D$ around, and predict using $p(x|D) = p(x, D)/p(D)$. Both terms can be computed exactly using DP, but this requires re-running DP for each new test case $x$. In addition, since the DP algorithm assumes complete data, if $x$ is incompletely observed (e.g., we want to "fill in" some of it), we must run the DP algorithm potentially an exponential number of times. For the same reason, we cannot sample from $p(x|D)$ using the DP method.

[1]For example, the number of DAGs on $d$ nodes for $d = 2 : 9$ are 3, 25, 543, 29281, 3781503, 1.1e9, 7.8e11, 1.2e15.

[2]Our Matlab/C implementation takes 1 second for $d = 10$ nodes and 6 minutes for $d = 20$ nodes on a standard laptop. The cost is dominated by the marginal likelihood computation, which all algorithms must perform. Our code is freely available at www.cs.ubc.ca/~murphyk/StructureLearning.

In this paper, we propose to fix all three of these shortcomings by combining DP with the Metropolis Hastings (MH) algorithm. The basic idea is simply to use the DP algorithm as an informative (data driven) proposal distribution for moving through DAG space, thereby getting the best of both worlds: a fast deterministic approximation, plus unbiased samples from the correct posterior, $G^s \sim p(G|D)$. These samples can then be used to compute the posterior mean of arbitrary features, $E[f|D] \approx \frac{1}{S}\sum_{s=1}^{S} f(G^s)$, or the posterior predictive distribution, $p(x|D) \approx \frac{1}{S}\sum_{s=1}^{S} p(x|G^s)$. Below we will show that this hybrid method produces much more accurate estimates than other approaches, given a comparable amount of compute time.

The idea of using deterministic algorithms as a proposal has been explored before e.g. [dFHSJR01], but not, as far as we know, in the context of graphical model structure learning. Further, in contrast to [dFHSJR01], our proposal is based on an exact algorithm rather than an approximate algorithm.

## 2 Previous work

The most common approach to estimating (features of) the posterior $p(G|D)$ is to use the Metropolis Hastings (MH) algorithm, using a proposal that randomly adds, deletes or reverses an edge; this has been called $MC^3$ for Markov Chain Monte Carlo Model Composition [MY95]. (See also [GC03] for some improvements, and [MR94] for a related approach called Occam's window [MR94].) Unfortunately, this proposal is very local, and the resulting chains do not mix well in more than about 10 dimensions. An alternative is to use Gibbs sampling on the adjacency matrix [MKTG06]. In our experience, this gets "stuck" even more easily, although this can be ameliorated somewhat by using multiple restarts, as we will see below.

A different approach, first proposed in [FK03], is to sample in the space of node orderings using MH, with a proposal that randomly swaps the ordering of nodes. For example,

$$(1, 2, 3, 4, 5, 6) \rightarrow (1, 5, 3, 4, 2, 6)$$

where we swapped 2 and 5. This is a smaller space ("only" $O(d!)$), and is "smoother", allowing chains to mix more easily. [FK03] provides experimental evidence that this approach gives much better results than MH in the space of DAGs with the standard add/ delete/ reverse proposal. Unfortunately, in order to use this method, one is forced to use a modular prior, which has various undesirable consequences that we discuss below (see Section 3). Ellis and Wong [EW06] realised this, and suggested using an importance sampling correction. However, computing the exact correction term is #P-hard, and we will see below that their approximate correction yields inferior results to our method.

An alternative to sampling orders is to analytically integrate them out using dynamic programming (DP) [KS04, Koi06]. We do not have space to explain the DP algorithm in detail, but the key idea is the following: when considering different variable orderings — say $(3, 2, 1)$ and $(2, 3, 1)$ — the contribution to the marginal likelihood for some nodes can be re-used. For example, $p(X_1|X_2, X_3)$ is the same as $p(X_1|X_3, X_2)$, since the order of the parents does not matter. By appropriately caching terms, one can devise an $O(d!2^d)$ algorithm to exactly compute the marginal likelihood and marginal posterior features.

To compute the posterior predictive density, $p(x|D)$, the standard approach is to use a plug-in estimate $p(x|D) \approx p(x|\hat{G}(D))$. Here $\hat{G}$ may be an approximate MAP estimate computed using local search [HGC95], or the MAP-optimal DAG which can be found by the recent algorithm of [SM06] (which unfortunately takes $O(d!2^d)$ time.) Alternatively, $\hat{G}$ could be a tree; this is a popular choice for density estimation since one can compute the optimal tree structure in $O(d^2 \log d)$ time [CL68, MJ00].

It can be proven that averaging over the uncertainty in $G$ will, on average, produce higher test-set predictive likelihoods [MGR95]. The DP algorithm can compute the marginal likelihood of the data, $p(D)$ (marginalizing over all DAGs), and hence can compute $p(x|D) = p(x, D)/p(D)$ by calling the algorithm twice. (We only need the "forwards pass" of [KS04], using the feature $f = 1$; we do not need the backwards pass of [Koi06].) However, this is very expensive, since we need to compute the local marginal likelihoods for every possible family on the expanded data set for every test case $x$. Below we will show that our method gives comparable predictive performance at a much lower cost, by averaging over a sample of graphs.

## 3 Modular priors

Some of the best current methods for Bayesian structure learning operate in the space of node orders rather than the space of DAGs, either using MCMC [FK03, EW06, HZ05] or dynamic programming [KS04, Koi06]. Rather than being able to define an arbitrary prior on graph structures $p(G)$, methods that work with orderings define a joint prior over graphs $G$ and orders $\prec$ as follows:

$$p(\prec, G) = \frac{1}{Z} \prod_{i=1}^{d} q_i(U_i^{\prec}) \rho_i(G_i) \times I(\text{consistent}(\prec, G))$$

where $U_i$ is the set of predecessors (possible parents) for node $i$ in $\prec$, and $G_i$ is the set of actual parents for node $i$. We say that a graph structure $G = (G_1, \ldots, G_d)$ is consistent with an order $(U_1, \ldots, U_d)$ if $G_i \subseteq U_i$ for all $i$. (In addition we require that $G$ be acyclic, so that $\prec$ exists.) Note that $U_i$ and $G_i$ are not independent. Thus the $q_i$ and $\rho_i$ terms can be thought of as factors or constraints, which define the joint prior $p(\prec, G)$. This is called a modular prior, since it decomposes into a product of local terms. It is important for computational reasons that $\rho_i(G_i)$ only give the prior weight to *sets* of parents, and not to their relative order, which is determined by $q_i(U_i)$.

From the joint prior, we can infer the marginal prior over graphs, $p(G) = \sum_{\prec} p(\prec, G)$. Unfortunately, this prior favors graphs that are consistent with more orderings. For example, the fully disconnected graph is the most probable under a modular prior, and trees are more probable than chains, even if they are Markov equivalent (e.g., 1←2→3 is more probable than 1→2→3). This can cause problems for structural discovery. To see this, suppose the sample size is very large, so the posterior concentrates its mass on a single Markov equivalence class. Unfortunately, the effects of the prior are not "washed out", since all graphs with the equivalence class have the same likelihood. Thus we may end up predicting that certain edges are present due to artefacts of our prior, which was merely chosen for technical convenience.

In the absence of prior knowledge, one may want to use a uniform prior over DAGs[3]. However, this cannot be encoded as a modular prior. To see this, let us use a uniform prior over orderings, $q_i(U_i) = 1$, so $p(\prec) = 1/(d!)$. This is reasonable since typically we do not have prior knowledge on the order. For the parent factors, let us use $\rho_i(G_i) = 1$; we call this a "modular flat" prior. However, this combination is not uniform over DAGs after we sum over orderings: see Figure 1. A more popular alternative (used in [KS04, Koi06, FK03, EW06]) is to take $\rho_i(G_i) \propto \binom{d-1}{|G_i|}^{-1}$; we call this the "Koivisto" prior. This prior says that different cardinalities of parents are considered to be equally likely a priori. However, the resulting $p(G)$ is even further from uniform: see Figure 1.

Ellis and Wong [EW06] recognized this problem, and tried to fix it as follows. Let $p^*(G) = \frac{1}{Z} \prod_i \rho_i^*(G)$ be the desired prior, and let $p(G)$ be the actual modular prior implied by using $\rho_i^*$ and $q_i = 1$. We can correct for the bias

[3]One could argue that we should use a uniform over PDAGs, but we will often be concerned with learning causal models from interventional data, in which case we have to use DAGs.

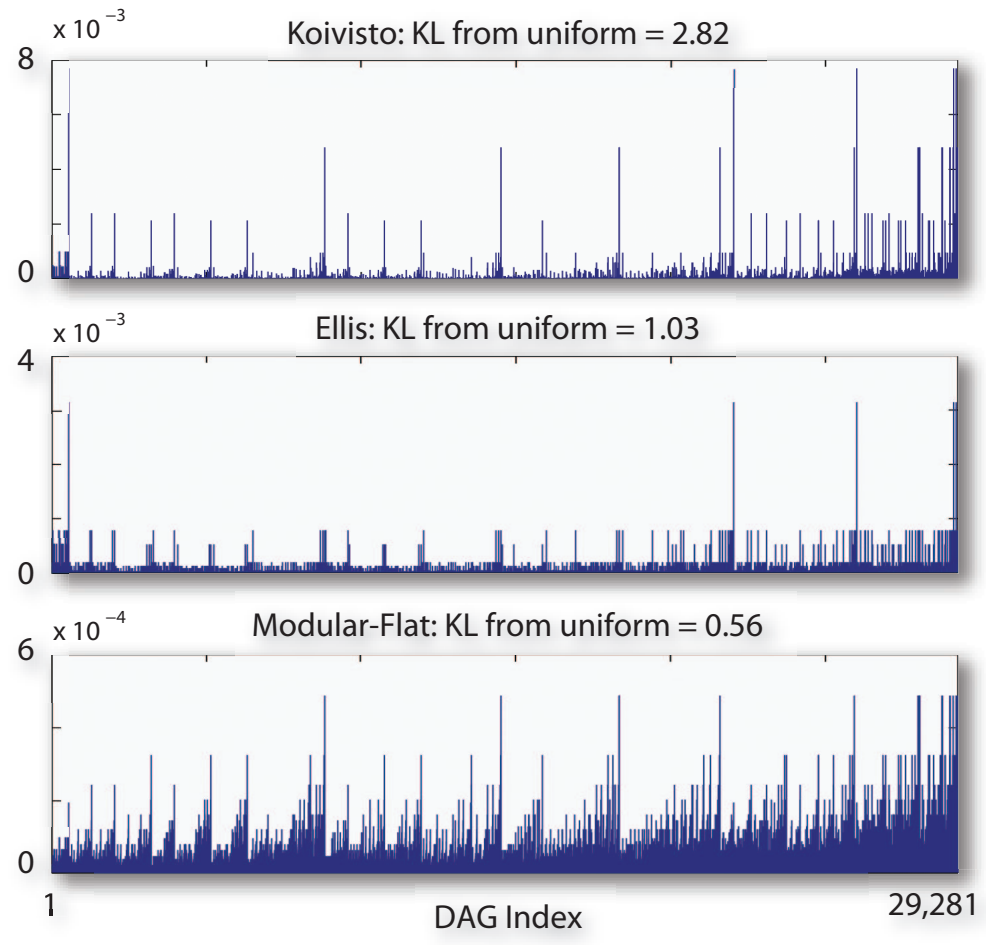

Figure 1: Some priors on all 29,281 DAGs on 5 nodes. Koivisto prior means using $\rho_i(G_i) \propto \binom{d-1}{|G_i|}^{-1}$. Ellis prior means the same $\rho_i$, but dividing by the number of consistent orderings for each graph (computed exactly). Modular flat means using $\rho_i(G_i) \propto 1$. This is the closest to uniform in terms of KL distance, but will still introduce artifacts. If we use $\rho_i(G_i) \propto 1$ and divide by the number of consistent orders. we will get a uniform distribution, but computing the number of consistent orderings is #P-hard.

by using an importance sampling weight given by

$$w(G) = \frac{p^*(G)}{p(G)} = \frac{\frac{1}{Z} \prod_i \rho_i^*(G_i)}{\sum_{\prec} \frac{1}{Z} \prod_i \rho_i^*(G_i) I(\text{consistent}(\prec, G))}$$

If we set $\rho_i^* = 1$ (the modular flat prior), then this becomes

$$w(G) = \frac{1}{\sum_{\prec} I(\text{consistent}(\prec, G)}$$

Thus this weighting term compensates for overcounting certain graphs, and induces a globally uniform prior, $p(G) \propto 1$. However, computing the denominator (the number of orders consistent with a graph) is #P-complete [BW91]. Ellis and Wong approximated this sum using the sampled orders, $w(G) \approx \frac{1}{\sum_{s=1}^{S} I(\text{consistent}(\prec^s, G))}$. However, these samples $\prec^s$ are drawn from the posterior $p(\prec |D)$, rather than the space of all orders, so this is not an unbiased estimate. Also, they used $\rho_i(G_i) \propto \binom{d-1}{|G_i|}^{-1}$, rather than $\rho_i = 1$, which still results in a highly non uniform prior, even after exact reweighting (see Figure 1). In contrast, our method can cheaply generate samples from an arbitrary prior.

## 4 Our method

As mentioned above, our method is to use the Metropolis-Hastings algorithm with a proposal distribution that is a mixture of the standard local proposal, that adds, deletes or reverses an edge at random, and a more global proposal that uses the output of the DP algorithm:

$$q(G'|G) = \begin{cases} q_{local}(G'|G) & \text{w.p. } \beta \\ q_{global}(G') & \text{w.p. } 1-\beta \end{cases}$$

The local proposal chooses uniformly at random from all legal single edge additions, deletions and reversals. Denote the set of acyclic neighbors generated in this way by $\text{nbd}(G)$. We have

$$q_{local}(G'|G) = \frac{1}{|\text{nbd}(G)|} I(G' \in \text{nbd}(G))$$

The global proposal includes an edge between $i$ and $j$ with probability $p_{ij}+p_{ji} \leq 1$, where $p_{ij} = p(G_{ij}|D)$ are the exact marginal posteriors computed using DP (using a modular prior). If this edge is included, it is oriented as $i{\rightarrow}j$ w.p. $q_{ij} = p_{ij}/(p_{ij}+p_{ji})$, otherwise it is oriented as $i{\leftarrow}j$. After sampling each edge pair, we check if the resulting graph is acyclic. (The acyclicity check can be done in amortized constant time using the ancestor matrix trick [GC03].) This leads to

$$q_{global}(G') = \left(\prod_i \prod_{j>i} (p_{ij}+p_{ji})^{I(G'_{ij}+G'_{ji}>0)}\right) \times \left(\prod_{ij} q_{ij}^{I(G'_{ij}=1)}\right) I(\text{acyclic}(G'))$$

We can then accept the proposed move with probability

$$\alpha = \min\left(1, \frac{p(D|G')p(G')}{p(D|G)p(G)} \frac{q(G|G')}{q(G'|G)}\right)$$

If we set $\beta = 1$, we get the standard local proposal. If we set $\beta = 0$, we get a purely global proposal. Note that $q_{global}(G')$ is independent of $G$, so this is an independence sampler. We tried various other settings of $\beta$ (including adapting it according to a fixed schedule), which results in performance somewhere in between purely local and purely global.

For $\beta > 0$ the chain is aperiodic and irreducible, since the local proposal has both properties [RC04, MY95]. However, if $\beta = 0$, the chain is not necessarily aperiodic and irreducibile, since the global proposal may set $p_{ij} = p_{ji} = 0$. This problem is easily solved by truncating edge marginals which are too close to 0 or 1, and making the appropriate changes to $q_{global}(G')$. Specifically, any $p_{ij} < C$ is set to $C$, while $p_{ij} > 1-C$ are set to $1-C$. We used $C = 1e-4$ in our experiments.

In this paper, we assume all the conditional probability distributions (CPDs) are multinomials (tables), $p(X_i = k|X_{G_i} = j, \theta) = \theta_{ijk}$. We make the usual assumptions of parameter independence and modularity [HGC95],

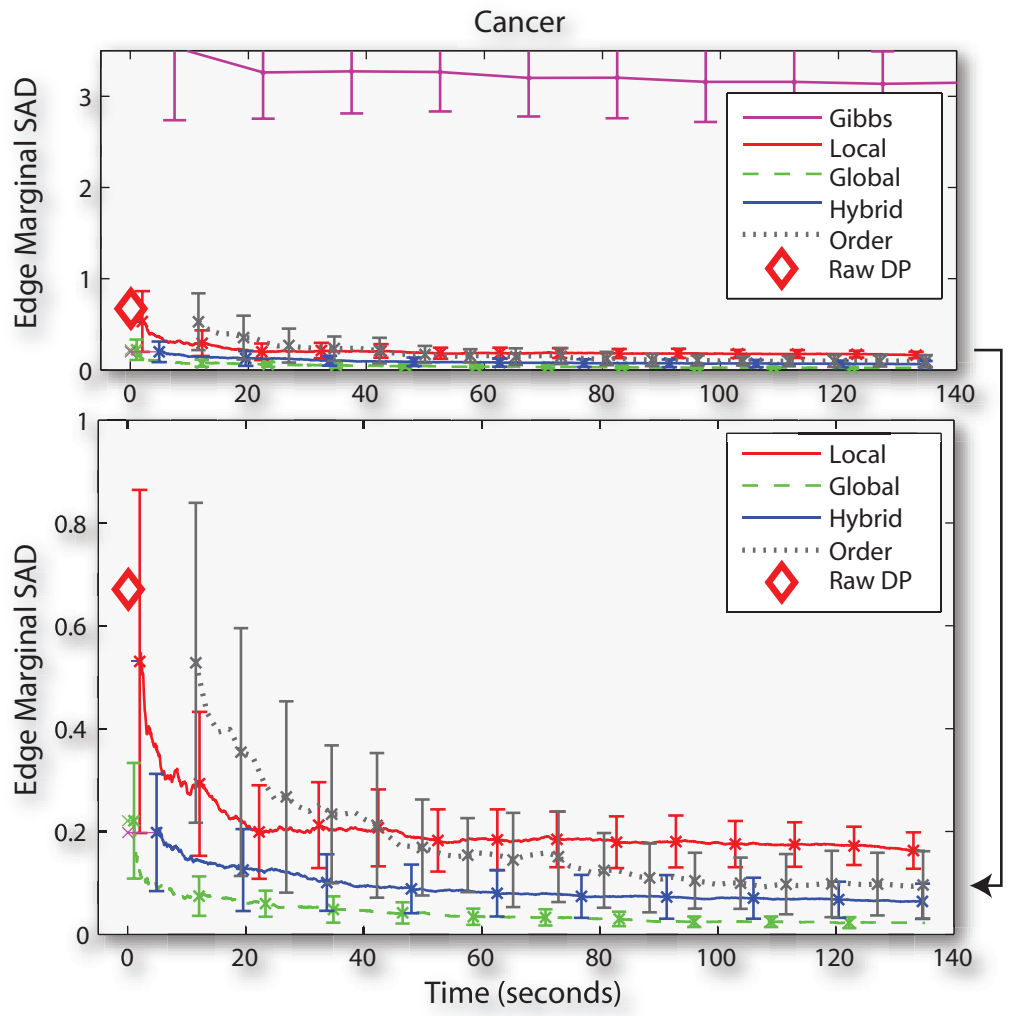

Figure 2: SAD error vs running time on the 5 node Cancer network. The Gibbs sampler performs poorly, therefore we replot the graph with it removed (bottom figure). Note that 140 seconds corresponds to about 130,000 samples from the hybrid sampler. The error bars (representing one standard deviation across 25 chains starting in different conditions) are initially large, because the chains have not burned in. This figure is best viewed in colour.

and we use uniform conjugate Dirichlet priors $\theta_{ij} \sim Dir(\alpha_i, \ldots, \alpha_i)$, where we set $\alpha_i = 1/(q_i r_i)$, where $q_i$ is the number of states for node $X_i$ and $r_i$ is the number of states for the parents $X_{G_i}$. The resulting marginal likelihood,

$$\begin{aligned} p(D|G) &= \prod_i p(X_i|X_{G_i}) \\ &= \prod_i \int [\prod_n p(X_{n,i}|X_{n,G_i}, \theta_i)] p(\theta_i|G_i) d\theta_i \end{aligned}$$

can be computed in closed form, and is called the BDeu (Bayesian Dirichlet likelihood equivalent uniform) score [HGC95]. We use AD trees [ML98] to compute these terms efficiently. Note that our technique can easily be extended to other CPDs, provided $p(X_i|X_{G_i})$ can be computed or approximated (e.g., using BIC).

# 5 Experimental results

## 5.1 Speed of convergence to the exact posterior marginals

In this section we compare the accuracy of different algorithms in estimating $p(G_{ij} = 1|D)$ as a function of their running time, where we use a uniform graph prior $p(G) \propto 1$. (Obviously we could use any other prior or feature of interest in order to assess convergence speed, but this seemed like a natural choice, and enables us to compare to the raw output of DP.) Specifically, we compute the sum of absolute differences (SAD), $S_t = \sum_{ij} |p(G_{ij} = 1|D) - q_t(G_{ij} = 1|D)|$, versus running time $t$, where $p(G_{ij} = 1|D)$ are the exact posterior edge marginals (computed using brute force enumeration over all DAGs) and $q_t(G_{ij}|D)$ is the approximation based on samples up to time $t$. We compare 5 MCMC methods: Gibbs sampling on elements of the adjacency matrix, purely local moves through DAG space ($\beta = 1$), purely global moves through DAG space using the DP proposal ($\beta = 0$, which is an independence sampler), a mixture of local and global (probability of local move is $\beta = 0.1$), and an MCMC order sampler [FK03] with Ellis' importance weighting term.[4] (In the figures, these are called as follows: $\beta = 1$ is "Local", $\beta = 0$ is "Global", $\beta = 0.1$ is "Hybrid".) In our implementation of the order sampler, we took care to implement the various caching schemes described in [FK03], to ensure a fair comparison. However, we did not use the sparse candidate algorithm or any other form of pruning.

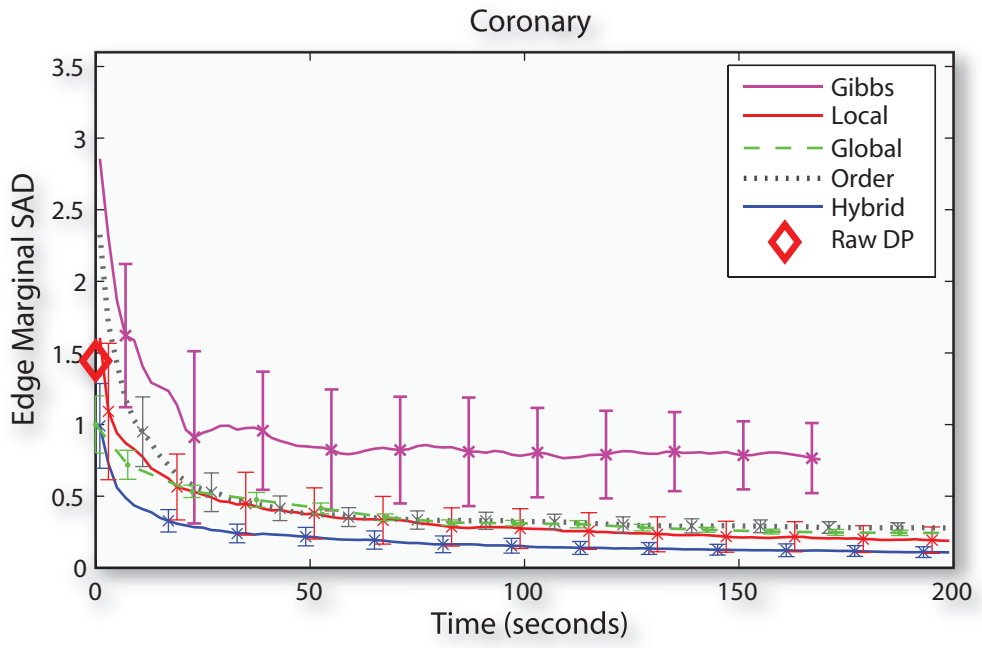

Figure 3: Similar to Figure 2, but on the 6 node CHD dataset.

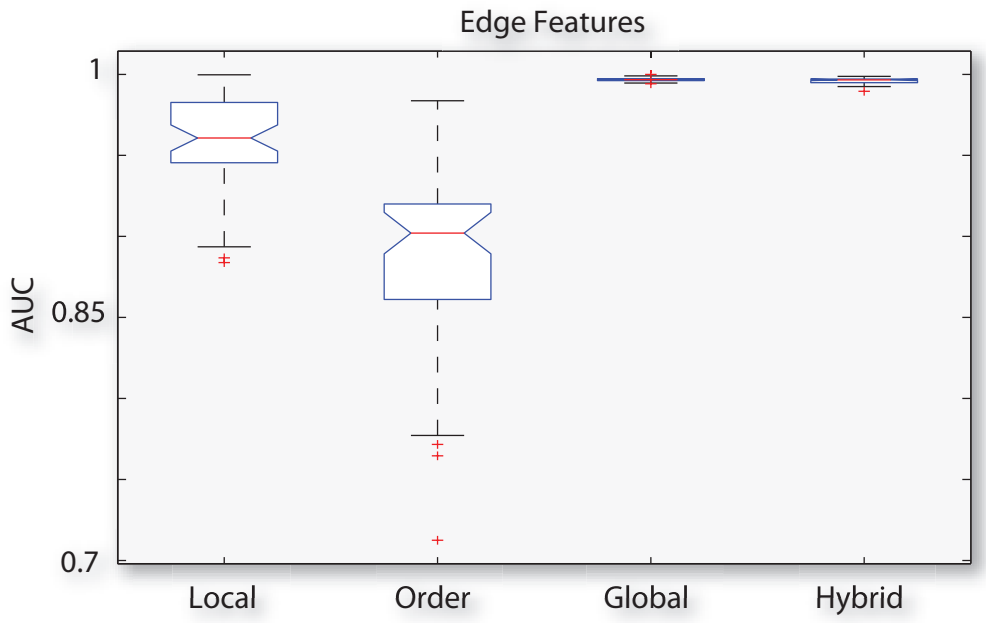

Figure 4: Area under the ROC curve (averaged over 10 MCMC runs) for detecting edge presence for different methods on the $d = 20$ node child network with $n = 10k$ samples using 200 seconds of compute time. The AUC of the exact DP algorithm is indistinguishable from the global method and hence is not shown.

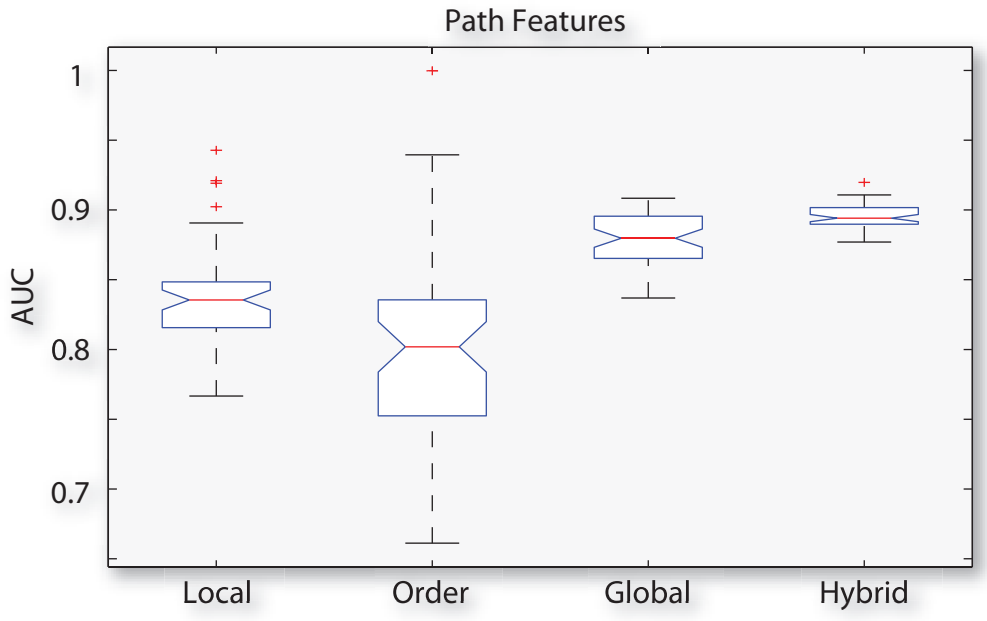

Figure 5: Area under the ROC curve (averaged over 10 MCMC runs) for detecting path presence for different methods on the $d = 20$ node child network with $n = 10k$ samples using 200 seconds of compute time.

For our first experiment, we sampled data from the 5 node "cancer network" of [FMR98] and then ran the different methods. In Figure 2, we see that the DP+MCMC samplers outperform the other samplers. We also consider the well-studied coronary heart disease (CHD) dataset [Edw00]. This consists of about 200 cases of 6 binary variables, encoding such things as "is your blood pressure high?", "do you smoke?", etc. In Figure 3, we see again that our DP+MCMC method is the fastest and the most accurate.

### 5.2 Structural discovery

In order to assess the scalability of our algorithm, we next looked at data generated from the 20 node "child" network used in [TBA06]. We sampled $n = 10,000$ records using random multinomial CPDs sampled from a Dirichlet, with hyper-parameters chosen by the method of [CM02] (which ensures strong dependencies between the nodes). We then compute the posterior over two kinds of features: edge features, $f_{ij} = 1$ if there is an edge between $i$ and $j$ (in either orientation), and path features, $f_{ij} = 1$ if there is a directed path from $i$ to $j$. (Note that the latter cannot be computed by DP; to compute it using the order sampler of [FK03] requires sampling DAGs given an order.) We can no longer compare the estimated posteriors to the exact posteriors (since $d = 20$), but we can compare them to the ground truth values from the generating network. Following [Hus03, Koi06], we threshold these posterior features at different levels, to trade off sensitivity and specifity. We summarize the resulting ROC curves in a single number, namely area under the curve (AUC).

The results for edge features are shown in Figure 4. We see that the DP+MCMC methods do very well at recovering the true undirected skeleton of the graph, obtaining an AUC of 1.0 (same as the exact DP method). We see that our DP+MCMC samplers are significantly better (at the 5% level) than the DAG sampler and the order sampler. The or-der sampler does not do as well as the others, for the same amount of run time, since each sample is more expensive to generate.

[4] Without the reweighting term, the MCMC order sampler [FK03] would give the same results (as measured by SAD) as the DP method [KS04, Koi06], only much, much slower.

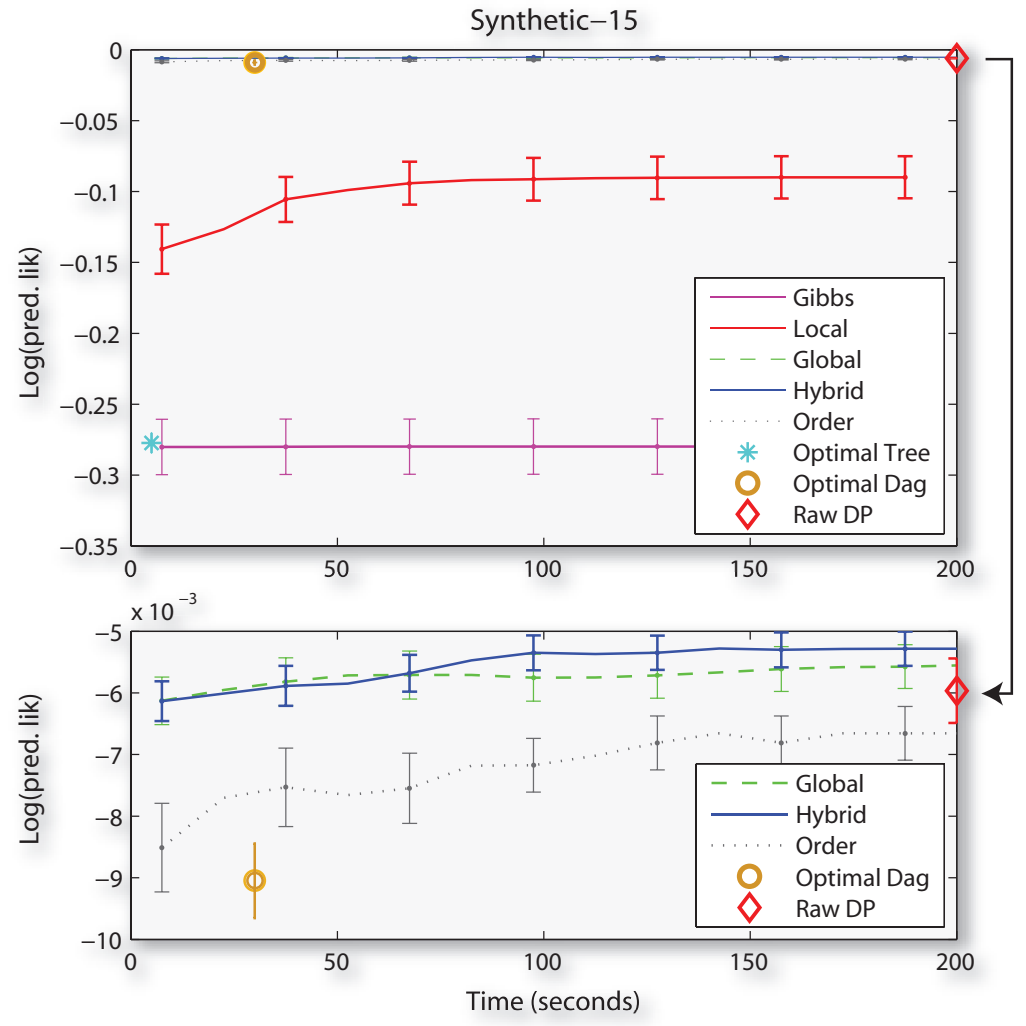

Figure 6: Test set likelihood vs training time on the synthetic $d = 15$, $N = 1500$ dataset. The bottom figure presents the "good" algorithms in higher detail by removing the poor performers. Results for the factored model are an order of magnitude worse and therefore not plotted. Note that the DP algorithm actually took over *two hours* to compute.

The results for path features are shown in Figure 5. Again we see that the DP+MCMC method (using either $\beta = 0$ or $\beta = 0.1$) yields statistically significant improvement (at the 5% level) in the AUC score over other MCMC methods on this much harder problem.

### 5.3 Accuracy of predictive density

In this section, we compare the different methods in terms of the log loss on a test set:

$$\ell = E \log p(x|D) \approx \frac{1}{m} \sum_{i=1}^{m} \log p(x_i|D)$$

where $m$ is the size of the test set and $D$ is the training set. This is the ultimate objective test of any density estimation technique, and can be applied to any dataset, even if the "ground truth" structure is not known. The hypothesis that we wish to test is that methods which estimate the posterior $p(G|D)$ more accurately will also perform better in terms of prediction. We test this hypothesis on three datasets: synthetic data from a 15-node network, the "adult" US census dataset from the UC Irvine repository and a biological dataset related to the human T-cell signalling pathway [SPP+05].

In addition to DP and the MCMC methods mentioned above, we also measured the performance of plug-in estimators consisting of: a fully factorized model (the disconnected graph), the maximum likelihood tree [CL68], and finally the MAP-optimal DAG gotten from the algorithm of [SM06]. We measure the likelihood of the test data as a function of *training time*, $\ell(t)$. That is, to compute each term in $\ell$ we use $p(x_i|D) = \frac{1}{S_t}\sum_{s=1}^{S_t} p(x_i|G^s)$, where $S_t$ is the number of samples that can be computed in $t$ seconds. Thus a method that mixes faster should produce better estimates. Note that, in the Dirichlet-multinomial case, we can quickly compute $p(x|G^s)$ by plugging in the posterior mean parameters:

$$p(x|G^s) = \prod_{ijk} \overline{\theta}_{ijks}^{I(x_i=j,x_{G_i}=k)}$$

where $\overline{\theta}_{ijks} = E[\theta_{ijk}|D, G^s]$. If we have missing data, we can use standard Bayes net inference algorithms to compute $p(x|G^s, \overline{\theta})$.

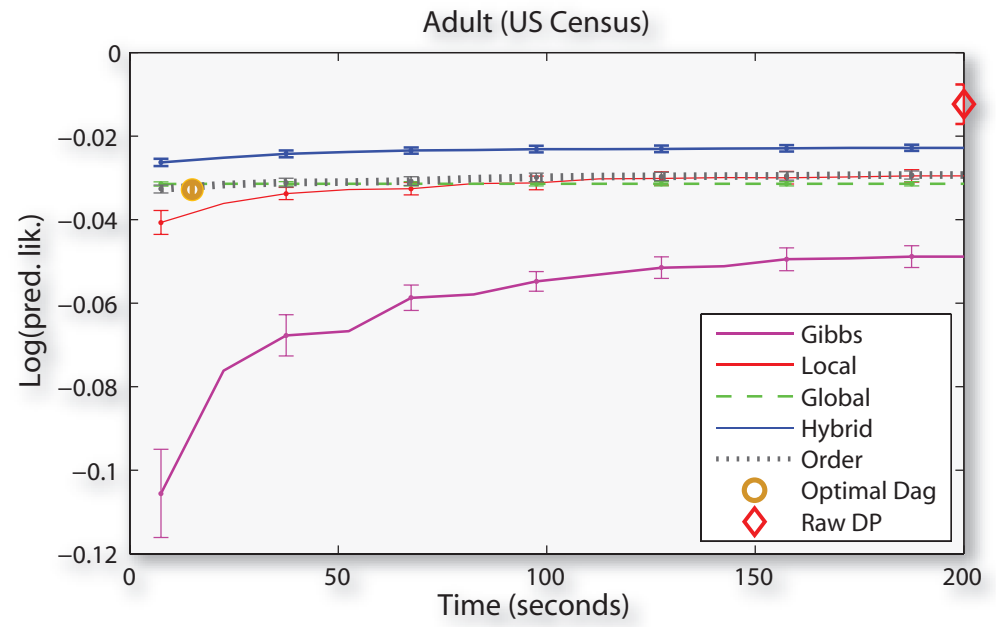

Figure 7: Test set log likelihood vs training time for the $d = 14, N = 49k$ "Adult" dataset. DP algorithm actually took over *350 hours* to compute. The factored and maximum likelihood tree results are omitted since they are many orders of magnitude worse and ruin the graph's vertical scale.

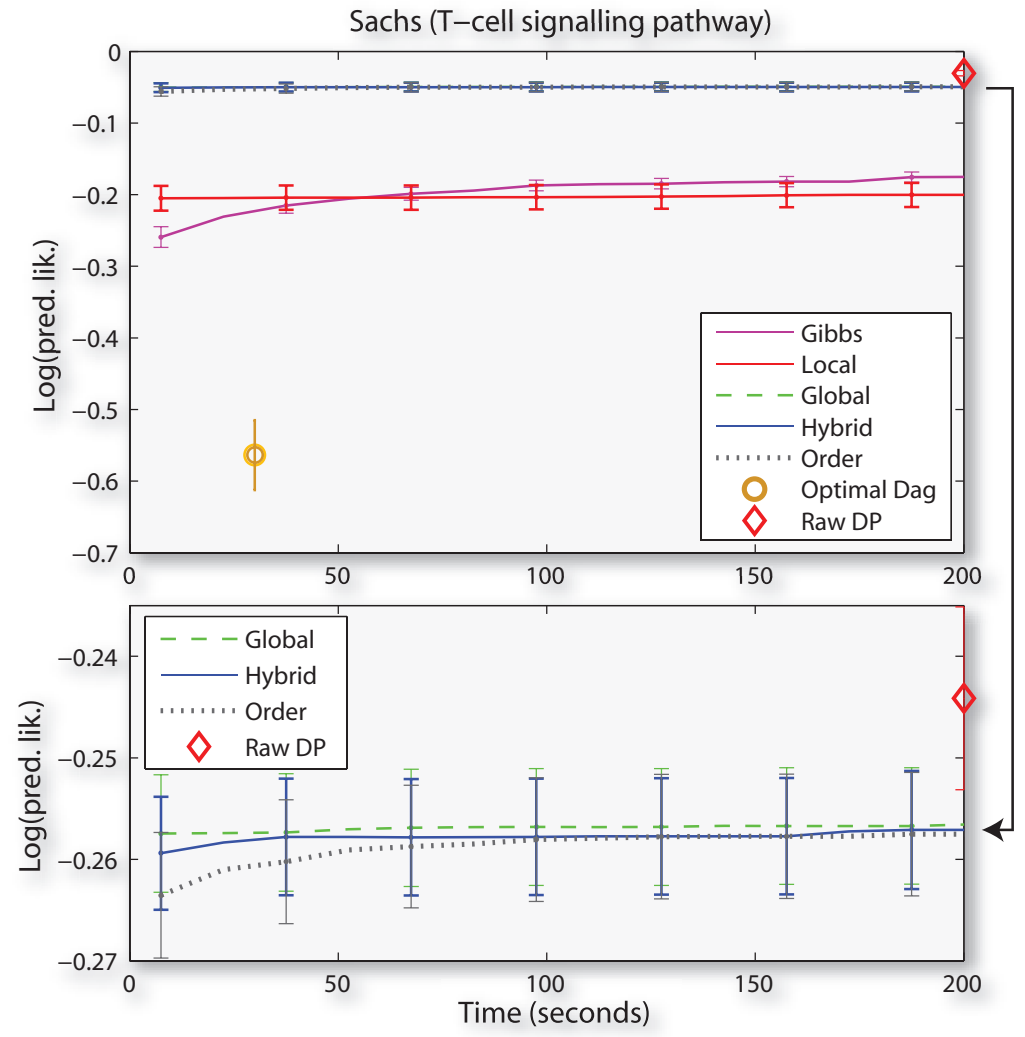

Figure 8: Test set log likelihood vs training time on the $d = 11, N = 5400$ T-cell data. DP algorithm actually took over *90 hours* to compute. The factored model and optimal tree plugins are again omitted for clarity. The bottom part of the figure is a zoom in of the best methods.

In contrast, for DP, the "training" cost is computing the normalizing constant $p(D)$, and the test time cost involves computing $p(x_i|D) = p(x_i, D)/p(D)$ for each test case $x_i$ separately. Hence we must run the DP algorithm $m$ times to compute $\ell$ (each time computing the marginal likelihoods for all families on the augmented data set $x_i, D$). DP is thus similar to a non-parametric method in that it must keep around all the training data, and is expensive to apply at run-time. This method becomes even slower if $x$ is missing components: suppose $k$ binary features are missing, then we have to call the algorithm $2^k$ times to compute $p(x|D)$.

For the first experiment, we generated several random networks, sampling the nodes' arities u.a.r. from between 2-4 and the parameters from a Dirichlet. Next, we sampled $100d$ records (where $d$ is the number of nodes) and performed 10-fold cross-validation. Here, we just show results for a 15-node network, which is representative of the other synthetic cases. Figure 6 plots the mean predictive likelihood across cross-validation folds and 5 independent sampler runs against training time. On the zoomed plot at the bottom, we can see that the hybrid and global MCMC methods are significantly better than order sampling. Furthermore, they seem to be better than exact DP, which is perhaps being hurt by its modular prior. All of these Bayes model averaging (BMA) methods (except Gibbs) significantly beat the plugin estimators, including the MAP-optimal structure.

In the next experiment we used the "adult" US census dataset, which consists of 49,000 records with 14 attributes, such as "education", "age", etc. We use the discretized version of this data as previously used in [MW03]. The average arity of the variables is 7.7. The results are shown in Figure 7. The most accurate method is DP, since it does exact BMA (although using the modular prior), but it is also the slowest. Our DP+MCMC method (with $\beta = 0.1$) provides a good approximation to this at a fraction of the cost (it took over 350 hours to compute the predictive likelihood using the DP algorithm). The other MH methods also do well, while Gibbs sampling does less well. The plug-in DAG is not as good as BMA, and the plug-in Chow-Liu tree and plug-in factored model do so poorly on this dataset that their results are not shown (lest they distort the scale). (These results are averaged over 10 MCMC runs and over 10 cross validation folds.)

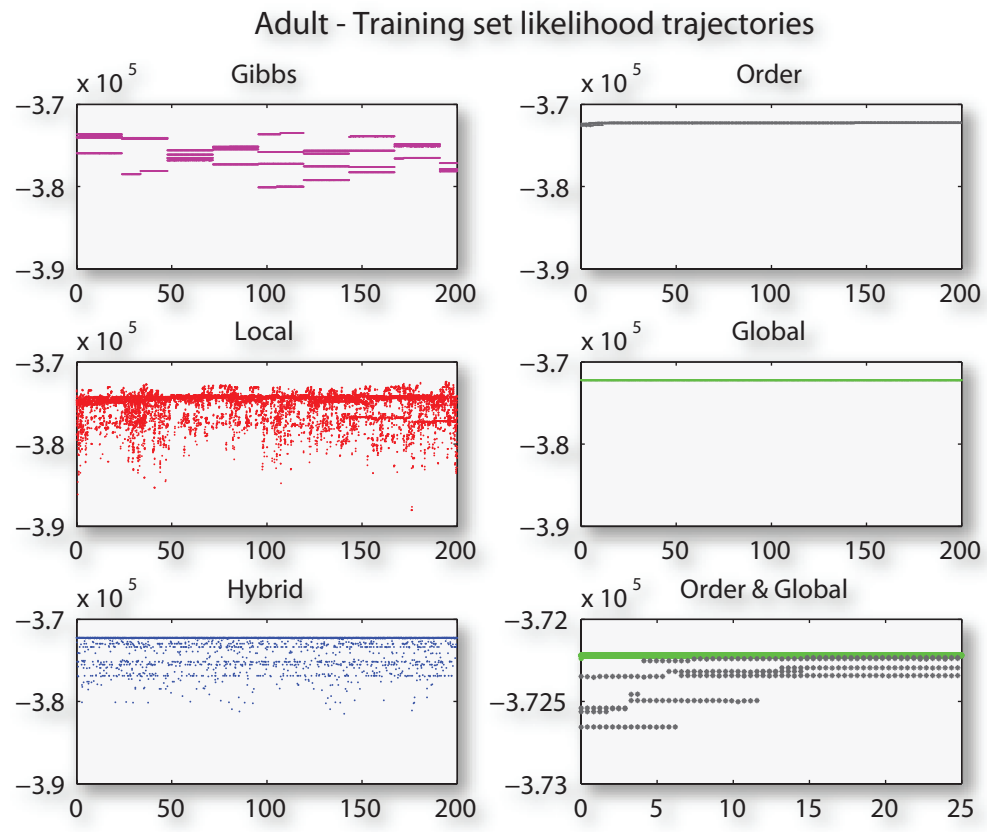

Figure 9: 4 traceplots of training set likelihood for each sampler on the adult dataset, starting from different values. The bottom-right figure combines runs from the order and global samplers and shows the behaviour of the chains in the first 25 seconds.

Finally, we applied the method to a biological data set [SPP+05] which consists of 11 protein concentration levels measured (using flow cytometry) under 6 different interventions, plus 3 unperturbed measurements. 600 measurements are taken in each condition yielding a total dataset of $N = 5400$ records. Sachs et al. discretized the data into 3 states, and we used this version of the data. We modified the marginal likelihood computations to take into account the interventional nature of the data as in [CY99]. The results are shown in Figure 8. Here we see that DP gives the best result, but takes 90 hours. The global, hybrid and order samplers all do almost at well at a fraction of the cost. The local proposal and Gibbs sampling perform about equally. All methods that perform BMA beat the optimal plugin.

### 5.4 Convergence diagonstics

In Figure 9 we show a traceplot of the training set marginal likelihood of the different methods on the Adult dataset. (Other datasets give similar results.) We see that Gibbs is "sticky", that the local proposal explores a lot of poor configurations, but that both the global and order sampler do well. In the bottom right we zoom in on the plots to illustrate that the global sampler is lower variance and higher quality than the order sampler. Although the difference does not seem that large, the other results in this paper suggest that the DP proposal does in fact outperform the order sampler.

## 6 Summary and future work

We have proposed a simple method for improving the convergence speed of MCMC samplers in the space of DAG models. Alternatively, our method may be seen as a way of overcoming some of the limitations of the DP algorithm of Koivisto and Sood [KS04, Koi06].

The logical next step is to attempt to scale the method beyond its current limit of 22 nodes, imposed by the exponential time and space complexity of the underlying DP algorithm. One way forward might be to sample partitions (layers) of the variables in a similar fashion to [MKTG06], but using our DP-based sampler rather than Gibbs sampling to explore the resulting partitioned spaces. Not only has the DP-based sampler been demonstrated to outperform Gibbs, but it is able to exploit layering very efficiently. In particular, if there are $d$ nodes, but the largest layer only has size $m$, then the DP algorithm only takes $O(d2^m)$ time. Using this trick, [KS04] was able to use DP to compute exact edge feature posteriors for $d = 100$ nodes (using a manual partition). In future work, we will try to simultaneously sample partitions and graphs given partitions.

## References

[BW91]G. Brightwell and P. Winkler. Computing linear extensions is #P-complete. In *STOC*, 1991.

[CL68]C. K. Chow and C. N. Liu. Approximating discrete probability distributions with dependence trees. *IEEE Trans. on Info. Theory*, 14:462–67, 1968.

[CM02]D. Chickering and C. Meek. Finding Optimal Bayesian Networks. In *UAI*, 2002.

[CY99]G. Cooper and C. Yoo. Causal discovery from a mixture of experimental and observational data. In *UAI*, 1999.

[DC04]D. Dash and G. Cooper. Model Averaging for Prediction with Discrete Bayesian Networks. *J. of Machine Learning Research*, 5:1177–1203, 2004.

[dFHSJR01]N. de Freitas, P. Hjen-Srensen, M. I. Jordan, and S. Russell. Variational MCMC. In *UAI*, 2001.

[Edw00]D. Edwards. *Introduction to graphical modelling*. Springer, 2000. 2nd edition.

[EW06]B. Ellis and W. Wong. Sampling Bayesian Networks quickly. In *Interface*, 2006.

[FK03]N. Friedman and D. Koller. Being Bayesian about Network Structure: A Bayesian Approach to Structure Discovery in Bayesian Networks. *Machine Learning*, 50:95–126, 2003.

[FMR98]N. Friedman, K. Murphy, and S. Russell. Learning the structure of dynamic probabilistic networks. In *UAI*, 1998.

[GC03]P. Giudici and R. Castelo. Improving Markov chain Monte Carlo model search for data mining. *Machine Learning*, 50(1–2):127 – 158, January 2003.

[HGC95]D. Heckerman, D. Geiger, and M. Chickering. Learning Bayesian networks: the combination of knowledge and statistical data. *Machine Learning*, 20(3):197–243, 1995.

[Hus03]D. Husmeier. Sensitivity and specificity of inferring genetic regulatory interactions from microarray experiments with dynamic Bayesian networks. *Bioinformatics*, 19:2271–2282, 2003.

[HZ05]K.-B. Hwang and B.-T. Zhang. Bayesian model averaging of Bayesian network classifiers over multiple node-orders: application to sparse datasets. *IEEE Trans. on Systems, Man and Cybernetics*, 35(6):1302–1310, 2005.

[Koi06]M. Koivisto. Advances in exact Bayesian structure discovery in Bayesian networks. In *UAI*, 2006.

[KS04]M. Koivisto and K. Sood. Exact Bayesian structure discovery in Bayesian networks. *J. of Machine Learning Research*, 5:549–573, 2004.

[MGR95]D. Madigan, J. Gavrin, and A. Raftery. Enhancing the predictive performance of Bayesian graphical models. *Communications in Statistics - Theory and Methods*, 24:2271–2292, 1995.

[MJ00]M. Meila and M. I. Jordan. Learning with mixtures of trees. *J. of Machine Learning Research*, 1:1–48, 2000.

[MJ06]M. Meila and T. Jaakkola. Tractable Bayesian learning of tree belief networks. *Statistics and Computing*, 16:77–92, 2006.

[MKTG06]V. Mansinghka, C. Kemp, J. Tenenbaum, and T. Griffiths. Structured priors for structure learning. In *UAI*, 2006.

[ML98]Andrew W. Moore and Mary S. Lee. Cached sufficient statistics for efficient machine learning with large datasets. *J. of AI Research*, 8:67–91, 1998.

[MR94]D. Madigan and A. Raftery. Model selection and accounting for model uncertainty in graphical models using Occam's window. *J. of the Am. Stat. Assoc.*, 89:1535–1546, 1994.

[MW03]Andrew Moore and Weng-Keen Wong. Optimal reinsertion: A new search operator for accelerated and more accurate bayesian network structure learning. In *Intl. Conf. on Machine Learning*, pages 552–559, 2003.

[MY95]D. Madigan and J. York. Bayesian graphical models for discrete data. *Intl. Statistical Review*, 63:215–232, 1995.

[RC04]C. Robert and G. Casella. *Monte Carlo Statisical Methods*. Springer, 2004. 2nd edition.

[Rob73]R. W. Robinson. Counting labeled acyclic digraphs. In F. Harary, editor, *New Directions in the Theory of Graphs*, pages 239–273. Academic Press, 1973.

[SM06]T. Silander and P. Myllmaki. A simple approach for finding the globally optimal Bayesian network structure. In *UAI*, 2006.

[SPP$^+$05]K. Sachs, O. Perez, D. Pe'er, D. Lauffenburger, and G. Nolan. Causal protein-signaling networks derived from multiparameter single-cell data. *Science*, 308, 2005.

[TBA06]I. Tsamardinos, L. Brown, and C. Aliferis. The max-min hill-climbing bayesian network structure learning algorithm. *Machine learning*, 2006.